\documentclass{article}
\usepackage{ijcai22}

\usepackage{amsmath,amsthm,amssymb}
\usepackage{algorithm}
\usepackage{algorithmic}
\usepackage{booktabs}
\usepackage[small]{caption}
\usepackage{diagbox}
\usepackage{datetime}
\usepackage{fancyhdr}
\usepackage{graphicx}
\usepackage[hidelinks]{hyperref}
\usepackage[utf8]{inputenc}
\usepackage{lastpage}
\usepackage[switch]{lineno} 
\usepackage{multirow}
\usepackage{soul}
\usepackage{subcaption}
\usepackage{times}
\usepackage{url}
\usepackage{xcolor}
\usepackage{etoolbox}
\apptocmd{\thebibliography}{\setlength{\itemsep}{0.5pt}}{}{}

\title{Entity Aware Modelling: A Survey}

\author{
Rahul Ghosh$^1$\and
Haoyu Yang$^1$\and
Ankush Khandelwal$^1$\and
Erhu He$^2$\and
Arvind Renganathan$^1$\and \\
Somya Sharma$^1$\and
Xiaowei Jia$^2$\And
Vipin Kumar $^1$
\affiliations
$^1$University of Minnesota, Twin Cities\\
$^2$University of Pittsburgh
\emails
\small
\{ghosh128, yang6993, khand035, renga016, sharm636, kumar001\}@umn.edu,
\{erh108, xiaowei\}@pitt.edu
}

\begin{document}

\maketitle

\begin{abstract}
    Personalized prediction of responses for individual entities caused by external drivers is vital across many disciplines. Recent machine learning (ML) advances have led to new state-of-the-art response prediction models. Models built at a population level often lead to sub-optimal performance in many personalized prediction settings due to heterogeneity in data across entities (tasks). In personalized prediction, the goal is to incorporate inherent characteristics of different entities to improve prediction performance. In this survey, we focus on the recent developments in the ML community for such entity-aware modeling approaches. ML algorithms often modulate the network using these entity characteristics when they are readily available. However, these entity characteristics are not readily available in many real-world scenarios, and different ML methods have been proposed to infer these characteristics from the data. In this survey, we have organized the current literature on entity-aware modeling based on the availability of these characteristics as well as the amount of training data. We highlight how recent innovations in other disciplines, such as uncertainty quantification, fairness, and knowledge-guided machine learning, can improve entity-aware modeling.
\end{abstract}


\vspace{-0.2in}
\section{Introduction}
\label{sec:introduction}
Personalized prediction is an essential task in many real-world applications, including recommendation systems~\cite{dhelim2022survey}, medical interventions~\cite{colberg2016physical}, and environmental sciences~\cite{ghosh2022robust}, which require robust personalized prediction models for sets of entities given limited training data for individual entities (or tasks). For example, the entities can represent a set of hydrological basins, and the objective is to model the streamflow response of several such basins for understanding hydrology cycles, water supply management, flood mapping, and reservoir operations. Similarly, in the healthcare domain, monitoring disease progression among patients (entity) or groups driven by external drivers, demographic or genetic information, and received treatments is essential for understanding the disease dynamics and downstream prediction task ~\cite{alaa2019attentive}. Other examples include personalized item prediction based on user behavior in e-Commerce systems~\cite{dhelim2022survey} or the forecasting of traffic patterns in different cities or countries~\cite{yao2019learning}. An entity can also be a physical system such as a drone or a spring-mass system where we want to model the trajectory of these systems.

\begin{figure}[t]
    \centering
    \includegraphics[width=0.45\linewidth]{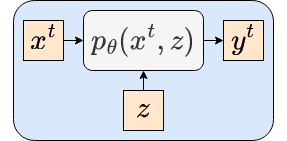}
    \caption{\textit{Forward model ($p_{\theta}$) which uses external drivers $(\boldsymbol{x^t})$ and entity characteristics $(\boldsymbol{z})$ to predict response $(\boldsymbol{y^t})$}}
    \label{fig:forward}
    \vspace{-0.2in}
\end{figure}

The major challenge in building personalized prediction models is the lack of training for individual entities. Hence, learning individual models can be sub-optimal, as shown by numerous studies in environmental~\cite{kratzert2019towards} and healthcare~\cite{tan2022metacare++,hussain2021neural}. On the other hand, a trivial merging of data from all entities to learn a single model will also fail to perform well. This is because the entity's response to the external drivers is governed by inherent properties specific to each entity. For example, for the same amount of precipitation (external driver), two river basins (entity) can have very different streamflow (response) values depending on their land-cover type and soil properties (entity characteristic)~\cite{newman2015gridded}. Similarly, in a clinical setting, a recorded treatment of medication (driver) for diabetes for a patient (entity) can have remarkably different effects (response) depending on the frequency of self-exercise (entity characteristic) \cite{colberg2016physical}. More examples of heterogeneity in entity characteristics include people (entity) having different heart rates (response) for the same physical activity (driver) depending on the physical fitness of each person (entity characteristic). Hence, ML methods must consider these entity characteristics to model the driver-response relationship effectively. We term this strategy of utilizing these entity characteristics to modulate the prediction model as entity-aware modeling (EAM). Figure~\ref{fig:forward} shows the diagrammatic representation of this EAM strategy. 

Various methods have been proposed across multiple disciplines to incorporate entity characteristics (implicitly or explicitly). The main issue with the current literature is that while the ideas developed across these disciplines apply to EAM, they have yet to be recognized or organized as such. This paper aims to provide an organized view of the literature around EAM, where the methods are borrowed from a wide range of applications and related ML tasks.

\begin{figure}[t]
    \centering
    \includegraphics[width=\linewidth]{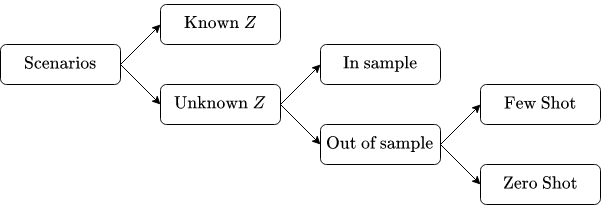}
    \caption{\textit{Problem Setting}}
    \label{fig:scenarios}
    \vspace{-0.1in}
\end{figure}

The critical goal of EAM is to build a global model that effectively leverages data from different entities by incorporating entity characteristics to reduce the impact of data scarcity for each entity. These techniques have been applied in several naturally occurring scenarios, as shown in Figure~\ref{fig:scenarios}. When entity-specific characteristics are explicitly available, they are often used directly in ML models for modulation~\cite{kim2022predictive,kratzert2019towards,liu2021metastore,logan2022patient,ghosh2022meta}. One advantage of having explicit characteristics (or learned embeddings) for each entity in the training set is that the learned model can be used for out-of-sample entities. However, these characteristics are often unknown or difficult to measure directly in many applications. Thus there is an additional need to build models that do not entirely depend on explicitly available characteristics. Such models either implicitly capture the entity/task characteristics as part of their parameter set~\cite{finn2017model} or infer entity/task embeddings from the data and use them to modulate the global network for personalization under heterogeneity~\cite{vuorio2019multimodal,suo2020tadanet,lin2021task}. Similarly, multi-task learning (MTL) framework \cite{taylor2017personalized,xia2020deepapp,wang2020human} can be used but at the expense of increased model complexity because MTL generally requires separate parameters for each task.

Due to the multi-faceted nature of EAM applications, innovations in other disciplines, such as the identifiability of entity characteristics and incorporating domain knowledge about entities, have a direct impact on improving the performance and usability of ML in these applications. Precisely, method advancements that can correctly identify the latent causal variables from the data~\cite{yang2021causalvae,shen2022weakly} can better characterize the entities. Similarly, several studies have shown that incorporating auxiliary information, either in the form of domain knowledge about the hierarchical structure~\cite{willard2022integrating} or additional observations about the state of entities, provides us with a way to monitor the evolving processes and characteristics of the entities. Furthermore, incorporating advances in the uncertainty quantification~\cite{garnelo2018neural,bishop1994mixture} and fairness~\cite{serna2020sensitiveloss} is not only pivotal for the usability of EAM in operational decision-making, but they can also lead to improved EAM.

To summarize, this survey aims to organize the diverse ML research threads proposed over the years that can be leveraged to tackle this EAM task. Furthermore, we enumerate the gaps and opportunities for advancing research in each direction. We organize the paper as follows. Section~\ref{sec:problem-formulation} first formulates the problem encountered in predicting the response for a diverse set of entities and discusses the different scenarios in this problem. Section~\ref{sec:methods} discusses the overarching themes between methods and applications. Lastly, Section~\ref{sec:furtherresearch} analyzes the additional topics that arise in this direction of EAM and lists open questions for future research.


\section{Entity-aware Prediction Scenarios}
\label{sec:problem-formulation}
This survey focuses on the setting where there are a set of $N$ entities/tasks. There exists a variability in the amount of training data available for entities - abundant, sparse or none. There may be a subset of well-observed entities, such that for each entity $i$ in this set, we have access to a training dataset $\mathcal{D}_i = \{(x_i^1, y_i^1), (x_i^2, y_i^2), \dots, (x_i^{T^{Train}_i}, y_i^{T^{Train}_i})\}$, with (drivers, response) pairs. From the remaining entities, there may be a subset of less-observed entities where, for each entity $j$ in the remaining set,  we have access to a few-shot dataset $\mathcal{D}_j = \{(x_j^1, y_j^2), \dots, (x_j^{T_{Few}}, y_j^{T_{Few}})\}$. The rest of the entities are completely unobserved.

The objective is to learn the mapping function from input variables $x_i^t$ to target variables $y_i^t$. In conventional supervised machine learning, we train a predictive model $\hat{y}_i^t = p_{\theta_i}(x_i^t)$, parameterized by $\theta_i$, by finding the parameters that minimize the empirical risk on the training data:
\begin{equation}
    \theta_i^* = \arg \min_{\theta_i} \mathcal{L}(\mathcal{D}_i;\theta_i)
\end{equation}
Given sufficient training data for each entity, we can train individual ML models that capture these inherent biases in each entity within the learned parameter set $\theta_i^*$. However, the data from all the entities are combined due to the lack of training data to learn a robust model for each entity. Learning a global model by the trivial merging of data from different entities can lead to sub-optimal results due to the heterogeneity across different sites. Because these entities are differentiated by their inherent characteristics $\boldsymbol{z}_i$, the functions are of the form $\hat{y}_i^t = p_{\theta}(x_i^t, \boldsymbol{z}_i)$, where $\theta$ denotes the function class shared by the target systems and $\boldsymbol{z}_i$ denotes entity-specific inherent characteristics as shown in Figure~\ref{fig:forward}.

Figure~\ref{fig:scenarios} summarizes the scenarios in which the ML models are used in real-world applications. The entity characteristics $\boldsymbol{z}_i$ can be explicitly available in some scenarios. In many scenarios, measurement of entity characteristics may be partially available for some of the characteristics, noisy or uncertain, or completely unavailable. In this situation, the entity characteristics must either implicitly be part of the models or be recovered from the data as latent variables. Further, the trained ML models can be evaluated in two further settings: (1) In-Sample test: the training and testing data are from the same entities but different from each other, and (2) Out-of-Sample test: the training and testing data are from different entities and different periods. In the out-of-sample testing scenario, we further have the few-shot and zero-shot setting depending on whether we have access to few-shot datasets for the testing entities. Next, we describe the most relevant literature for these scenarios.


\section{Methods}
\label{sec:methods}

\subsection{Known $\boldsymbol{z}$}
In the scenario where entity-specific characteristics are known, the most standard approach is to modulate the network using them~\cite{kim2022predictive,logan2022patient}. Several studies have used these source characteristics by concatenating them along with the drivers and then passing them into the ML models~\cite{kratzert2019towards}. Other approaches have used entity-specific characteristics/features to modulate the architecture or transform the input features. In the context of store placement prediction in multiple cities, \cite{liu2021metastore} use the city-specific parameters in an attention network to modulate and adapt the base feature extractor. Further, \cite{perez2018film} perform feature-wise affine transformation based modulation using entity characteristics based conditioning. Similarly, \cite{wang2021dcn} explicitly model the interaction of the input drivers and entity characteristics through a deep and cross network. One advantage of having explicit characteristics for each entity is that the learned model can be used for out-of-sample entities. Thus we do not further divide this scenario based on whether the ML models are being applied in the in-sample or out-of-sample setting, which is trivial. However, all the methods discussed in the later settings can be easily adapted for this scenario. A challenge commonly faced in this scenario is handling the model's bias towards certain types of entities caused due to the fact that the training set may be imbalanced in the types and occurrences of entities, as discussed in Section~\ref{sec:fairness}.

\subsection{Unknown $\boldsymbol{z}$ \& In-sample}
When the characteristics are unavailable, EAM methods that learn to leverage entity relationship implicitly is required. \textbf{Multi-task Learning} (MTL) is the most common approach used for several personalized prediction applications in this setup, such as mood prediction~\cite{taylor2017personalized}, app-usage prediction~\cite{xia2020deepapp}, and human mobility prediction~\cite{wang2020human}. The different tasks/entities (used interchangeably here) share a common network in multi-task learning followed by task-specific weights to achieve personalization. However, there are two main challenges with MTL for personalized prediction. First, the number of entity-specific weights increases rapidly with the number of entities. Furthermore, the shared network should have sufficient capacity to handle a large set of entities~\cite{wang2021bridging}. Hierarchical Dirichlet processes have been used to combine similar tasks at the expense of the increased model and computational complexity. Recently, \cite{liu2020multi} proposed a deep MTL framework that aims to learn entity similarity from the data to reduce the impact of limited training data while training entity-specific parameters. Another approach in this scenario is to train global models by assigning one-hot/random vector to each entity~\cite{li2022regionalization}. Here the entity-specific parameters do not depend on the number of entity but on the dimensionality of random characteristics, thus reducing the high model complexity of the MTL framework. More complex methods, such as meta-learning (discussed later), can also be applied in this scenario.

\subsection{Unknown $\boldsymbol{z}$, out-of-sample \& few-shot}
When few samples of observation are available for the out-of-sample entities, few-shot learning~\cite{wang2020generalizing} methods can be used. MTL methods are not the right approach as a model trained using traditional learning schemes is not easily adaptable to a different set of  entities~\cite{finn2017model}. The solution is to use the few-shot data to either adapt the models to the new entities through gradient-based optimization or infer the entity characteristics and use them to modulate the prediction model.

\noindent\textbf{Meta Learning:} Recently meta-learning has gained much attention in few-shot learning applications by leveraging the shared structure between existing training tasks, leading to better generalization and adaptation~\cite{hospedales2021meta}. In particular, Model Agnostic Meta Learning (MAML)~\cite{finn2017model} aims to learn a global meta model that can be easily adapted to create personalized models for each entity. This is commonly done by formulating the training scheme as a bi-level optimization problem:
\begin{equation}
    \begin{split}
        \theta^* &= \arg \min_{\theta} \sum_{i\in \mathcal{P}(i)} \mathcal{L}(\mathcal{D}^{val}_i;\theta_i^*)\\
        \text{s.t.} \qquad \theta^*_i &= \arg \min_{\theta} \mathcal{L}(\mathcal{D}^{train}_i;\theta)
    \end{split}
\end{equation}

During meta-training, the individual models $\theta_i$ are finetuned for each entities using their meta-training samples $\mathcal{D}^{train}_i$. These individual models are used to calculate the loss on meta-test samples $\mathcal{D}^{val}_i$, which serves as the training error for the meta model $\theta^*$. This meta-model can adapt to each entity using one or a small number of gradient steps to find the task-adapted parameter of the prediction model.

Meta-learning has gained much attention in recent years for several EAM tasks. \cite{lee2021individualized} obtain a high-performance personalized model using meta learning and few-shot entity data. \cite{tan2022metacare++} bridge the modeling of infrequent patients (entites) and rare diseases (tasks) by designing a meta learning approach based on hierarchical patient subtyping mechanism. \cite{ajila2022multilayer} show the benefit of meta-learning over individual models in forecasting a diverse set of air pollution. \cite{zhang2019metapred} developed a MAML framework for multiple clinical risks prediction in healthcare application. Other applications include using the prior consumption data from multiple source cities to predict optimal store placement in a new city~\cite{liu2021metastore}.

Adapting the whole parameter set may put extensive burden on the optimization procedure, possibly biasing the solution of the inner-level optimization. Recently, variations of MAML have been proposed that adapt only the high-level layers instead of the whole meta-model~\cite{raghu2019rapid,zhao2019learning}. This strategy can be adapted for the EAM modeling. The key idea is to freeze the prediction model in the inner loop and assume only entity characteristics as a trainable vector. This strategy has been used in engineering~\cite{kim2022predictive}, finance~\cite{liu2020calibration}, and vision domains~\cite{zintgraf2019fast}. Recently, \cite{tayal2022invertibility} used an invertible neural network to infer lake attributes using a few observations. Similarly, \cite{park2022meta} proposed to jointly identify and predict systems using similar bi-level optimization of MAML:
\begin{equation}
    \begin{split}
        \theta^* &= \arg \min_{\theta} \sum_{i\in \mathcal{P}(i)} \mathcal{L}(\mathcal{D}^{val}_i;\theta,z_i^*)\\
        \text{s.t.} \qquad z^*_i &= \arg \min_{z} \mathcal{L}(\mathcal{D}^{train}_i;\theta, z)
    \end{split}
\end{equation}
Here, the total parameters are separated into prediction model $\theta$ shared by the target entities and entity-specific characteristics $z$. Additionally, many MAML-based methods assume that all train and test entities are drawn from the same distribution. Thus, a single meta-initialization could be challenging to adapt due to the data distribution in different entities being different and multimodal~\cite{vuorio2019multimodal}. Furthermore, the training process is computationally expensive and sensitive to hyperparameter choices~\cite{antoniou2018train}.

\noindent\textbf{Conditional Meta Learning:} If the entity distribution is multi-modal with disjoint and far apart modes (e.g. patients/groups from different countries), a set of separate meta-learners could better master the full distribution. Several strategies have been proposed to learn meta-learners that acquire mode-specific prior parameters by formulating the bi-level optimization problem as,
\vspace{-0.07in}
\begin{equation}
    \begin{split}
        \theta^* &= \arg \min_{\theta} \sum_{i\in \mathcal{P}(i)} \mathcal{L}(\mathcal{D}^{val}_i;\theta_i^*)\\
        \text{s.t.} \quad \theta^*_i &= \arg \min_{\theta} \mathcal{L}(\mathcal{D}^{train}_i;\theta)\quad\text{\textit{and}}\quad\theta = \mathcal{T}(\boldsymbol{z})
    \end{split}
\end{equation}

Here, the meta-model is conditioned on additional side information $\mathcal{T}(\boldsymbol{z})$ that contains descriptive features associated to the entity/task. Several works have been proposed that advocate this conditional perspective. They have been called several names such as heterogeneous meta learning~\cite{chen2021hetmaml}, conditional meta learning~\cite{denevi2021conditional} or multi-modal meta learning~\cite{vuorio2019multimodal}. 

Associating each entity with one of the meta initializations require additional entity characteristics, which is often unavailable or could be ambiguous when the modes are not disjoint. Under this setting, the most common strategy is to learn another network that converts training data from seen entities into entity-specific embeddings that modulate the shared prediction network~\cite{vuorio2019multimodal}. The prediction and embedding networks can be trained either jointly or alternately. \cite{suo2020tadanet} learn embedding metric space that characterizes disease (entity) relationships for disease prediction and shows promising results for solving the data scarcity problem in healthcare decision support. \cite{jerfel2019reconciling} learn a mixture of hierarchical Bayesian models by incorporating entity-specific parameters as latent variables. This allows the meta-learner to perform entity-specific parameter selection instead of consolidating inductive biases into a single meta-model. Further, \cite{yao2019learning} showed that utilizing the traffic data from data-rich cities improves the prediction in cities with only a short period of data using a conditioned MAML (CMAML) based approach. Similarly, \cite{chen2022physics} proposed a graph-based conditional meta-learning approach for predicting water quality and quantity variables in a diverse set of basins. One challenge of CMAML that needs to be addressed is quantifying the diversity needed to merit these methods. Further, most of CMAML methods utilize a metric that measures the similarity between entities. Thus we need to investigate novel metrics that better capture this similarity.

\noindent\textbf{Neural Process:} The neural process (NP) family has been used in EAM in a variety of fields, including robotics, computer vision, and natural language processing~\cite{jha2022neural}. The Neural Process (NP) family of methods started with Conditional Neural Processes (CNPs)~\cite{garnelo2018conditional}, which combine the benefits of Deep neural networks and Bayesian methods, such as Gaussian Processes (GPs), to exploit prior knowledge and quickly infer the shape of a new function. The defining characteristic of the NP framework is that it conditions the prediction function on the observations via an inferred entity embedding. The resulting model can be boiled down to three core components, as shown below,
\begin{linenomath}\begin{align}
    h_c &= q_{\phi}(x_c, y_c) \qquad && \text{encoder} \notag\\
    z &= h_1 \oplus \dots \oplus h_n \qquad && \text{aggregator}\\
    y_c &= p_{\theta}(x_c, z) \qquad && \text{conditional decoder} \notag
\end{align}\end{linenomath}

Here, the encoder produces a representation from each (input, output) pair, that are aggregated to form an embedding. The conditional decoder outputs the target predictions using the embedding and inputs. Further advancements have been proposed, such as introduction of latent variables~\cite{garnelo2018neural} instead of deterministic embeddings, using bootstrapping for multiple latent variables~\cite{lee2020bootstrapping}, or introducing attention-based versions~\cite{kim2019attentive}.

The NP framework has found applications in a wide range of domains, given its flexibility, modeling capacity, and computational efficiency. NPs have been used in designing recommender systems~\cite{lin2021task} for personalized prediction of an item for each user (entity). In neuroscience, NPs have been used to predict the responses of neurons (entities) in the visual cortex to natural stimuli~\cite{cotton2020factorized} and neural spike sorting~\cite{pakman2020neural}. \cite{wu2022multi} propose a Multi-fidelity Hierarchical Neural Process (MF-HNP) that can leverage the cheap data from low-fidelity simulators for epidemiology tasks across individuals from multiple age groups and climate modeling for diverse sites. \cite{requeima_fast_2019} propose an extension to the NP framework for multi-task classification settings that can quickly adapt to a new task without costly retraining. \cite{ghosh2022robust} use a self-supervised contrastive loss to infer the entity embeddings for personalized streamflow prediction.

Despite the success of the NP framework, there exist open challenges and limitations that need to be addressed. Deciding the aggregator function is still an open direction of research~\cite{yoon2020robustifying}. Further in many scientific process there are multiple processes within an entity which can lead to multiple contexts, as discussed in Sec~\ref{sec:aux}. Thus a potential direction is to impose a manifold structure on the latent distribution or a hierarchy among the latent distributions from the diverse contexts for the same entity.


\subsection{Unknown $\boldsymbol{z}$, out-of-sample \& zero-shot}
In many scenarios, a good model is expected for these out-of-sample entities, despite collecting high-quality data for all possible entities (e.g., abnormalities/diseases in healthcare) being challenging. While meta-learning is the common approach for few-shot learning scenarios, it cannot be used when we have no data available for out-of-sample entities (zero-shot setting). Since there is no data/knowledge about the entity characteristics, the characteristics should only be inferred from the training entities' drivers and responses. There could be multiple choices for the latent characteristics that can yield the same data distribution. However, only one (or a small subset) contributes a robust model. Additionally, the global model could be biased toward the in-sample entities.

\noindent\textbf{Disentangled Representation Learning:} Disentangled representation learning is proposed to partition the hidden representation $h$ into independent factors of variations, which are aligned with data generative factors~\cite{locatello2019challenging}. For example, in the image classification task, a disentangled representation might encode the shape and color of an object separately. Based on generative models like variational autoencoders (VAEs) and generative adversarial networks (GANs) structure, disentanglement is encouraged by new regularizations and training techniques~\cite{chen2016infogan,Higgins2017betaVAELB}. Better disentanglement could be achieved if there is more information about the latent generative factors, like hierarchical priors or a group of entities sharing a common factor~\cite{deng2017factorized,bouchacourt2018multi}. To build an entity-aware model using disentangled representations, one option is to separate the entity-dependent representations from the representations which are shared by all the entities, i.e., $h_i = [h_{shared}, z_i]$. Recent progress in disentangled representation learning provides opportunities for this approach. For example, using identification label, \cite{eom2019learning} introduced identity shuffle GAN (IS-GAN) to disentangle identity-related (e.g., clothing) and unrelated features (e.g., human pose) from person images. \cite{yingzhen2018disentangled} introduced a disentangled sequential auto-encoder. The latent representation is learned to separate time-independent (e.g., static entity characteristics) and time-dependent features (e.g., states of the entity).

\noindent\textbf{State Space Model:} State space model (SSM) is a model designed for sequential data, which assumes the observational data is generated from latent state variables through \textit{emission model}. The transitions between latent states are modeled by \textit{transition model}. Given observations $[x^t, y^t]$ and latent states $h^t$, the vanilla state space model~\cite{durbin2012time} can be formulated as 
\begin{linenomath}\begin{align}
    \label{eq:state_space_model}
    h^t &= g(h^{t-1}) + \varepsilon_z &\qquad\text{\textit{(transition model)}}\notag\\
    [x^t, y^t] &= f(h^{t}) + \varepsilon_x &\qquad\text{\textit{(emission model)}}
\end{align}\end{linenomath}

According to the equation above, the hidden Markov model (HMM) can be seen as a special state space model, where the latent state is discrete and the transition only depends only on the previous latent state. Recently, researchers added neural structures to the conventional state space model for better approximation and to learn nonlinear latent states. The deep SSMs are usually solved by variational learning algorithm, which includes a inference network to approximate the intractable posterior of latent states and a generative model to approximate the transition and emission model. \cite{krishnan2017structured} proposed an inference algorithm to learn continuous latent states of deep Markov models (DMMs), where the emission distributions are modeled by deep neural networks, and the transition distributions are estimated by an RNN-based inference network. \cite{alaa2019attentive} put attention mechanism on latent states to investigate the dependence between the current and all past states, which generalize the transition model in Eq.\ref{eq:state_space_model}. The authors also proposed an inference algorithm for a discrete latent state. \cite{tang2021probabilistic} further combined state space model with transformer architectures, which uses attention mechanism instead of RNNs to model latent state dynamics. Given its flexibility and interpretability, the state space model is widely used for time series modeling and forecasting in different domains like computer vision~\cite{fraccaro2017disentangled}, and healthcare~\cite{xue2020deep}. Recent progress in the state space model provides great promise for EAM. The overall idea is to use prior knowledge about entities to learn better latent state representations and dynamics. For example, \cite{hussain2021neural} introduced a model that allows the transition of latent states depends on the characteristics of entity (e.g., genetics, demographics). The learned latent representations thus implicitly captures the entity-related knowledge (e.g., clinical phenotypes, pharmacodynamic) from the observations. In computer vision domain, \cite{fraccaro2017disentangled} used a Kalman variational auto-encoder (KVAE) as inference network. The KVAE is designed to separate the object’s representation from latent state describing its dynamics in a unsupervised manner, which overlaps with disentangled representation.

\noindent\textbf{Causal Representation Learning:} This group of methods focus on the discovery of latent causal variables and the robust prediction in the downstream task~\cite{scholkopf2021toward}. Most disentangled representation learning methods are insufficient to learn causal representations since they try to disentangle independent factors from observations. However, causal factors are usually dependent on each other, which forms an underlying causal structure. To fill this gap, a line of recent work focuses on recovering the causal representation from the disentangled factors. \cite{yang2021causalvae} proposed to add a causal layer in the VAE-based model to transform independent factors into causal representation. \cite{reddy2022causally} introduced a weakly supervised disentanglement method when the dependency among hidden generative factors is only caused by confounders (common parents). \cite{shen2022weakly} used a trainable structural causal model as the prior distribution to enforce causal disentanglement, instead of an independent one. In causal representation learning, the \textit{Causality Assumption}~\cite{buhlmann2020invariance} states that the environment $e$ does not change the relationship between covariates $X$ and target variables $Y$.The environment $e \in \mathcal{E}$ is a special case of an entity, which is also referred to as experimental setting, sub-population, or perturbation. For example, basins from different locations, and different patient populations can be seen as different environments. Leveraging such invariance across entities could yield a robust model. Given the assumption that environment $e$ only change the distribution of covariates $X$, recent progress show empirically and theoretically that causal representations enable out-of-distribution generalization \cite{lu2021invariant}.

Entity-aware models built on causal representation can resist the distributional shifts induced by interventions, and selection bias. However, this approach may fail when entities have divergences other than distributional shifts. Further, domain knowledge on latent causal variables/mechanisms could be critical for the causal identification~\cite{wangsound}. Successful adoption of the causal representation learning methods require addressing the challenge of the identifiability of causal variables~\cite{locatello2019challenging}. The detailed discussion of the problem of identifiability can be found in section~\ref{sec:Identifiability}.


\section{Further research topics}
\label{sec:furtherresearch}

\subsection{Incorporating additional entity level information}
\label{sec:aux}
Several applications exist where auxiliary information about entities can be accessed. This supplementary information can be available in primarily two forms: a) process understanding of the entities and b) additional independent observations of entity states. ML models, being data-driven, are not impacted by our limited understanding of the underlying processes. However, ML models can only learn (however complex) patterns in the data used for training and thus fail on unseen data that is outside the range seen in training. Most real-world systems consist of multiple physical processes interacting in a hierarchical order. Moreover, these processes are often highly nonlinear and exhibit complex behavior encompassing multiple inputs and outputs. There is an opportunity to advance the EAM framework further by leveraging prior physical knowledge of the hierarchical structure. The hierarchical structure provides a principled way to share parts of the entity characteristics across diverse processes through joint optimization. Apart from advances in physics-guided machine learning that utilize physical equations, boundary conditions, and other inductive biases, entity-specific physical descriptors and physical processes can also be incorporated in modeling framework to enable generalization in unseen scenarios~\cite{willard2022integrating}. Another opportunity unique to many environmental problems is the availability of ancillary information about the system beyond the standard input and output variables. For example, streamflow in a river catchment is modeled as a function of weather drivers, but auxiliary information such as soil moisture data from in-situ sensors or earth observing satellites~\cite{entekhabi2010soil} can provide valuable information related to underlying processes such as evapotranspiration and base flow. New EAM methods are required that can readily incorporate such diverse sources of data and has the potential to represent complex physical relationships between multiple bio-geo-physical processes.

\subsection{Identifiability of Characteristics/Equifinality}
\label{sec:Identifiability}
When characteristics are unknown in EAM, a central problem is how to correctly identify those factors. Although methods like NP, SSM, and disentangled/causal representation learning show potential to learn entity-related representations, there is no guarantee the learned latent representation corresponds to the real characteristics (latent causal factors)~\cite{locatello2019challenging}. The intuition is that given observational variables, there could be infinitely many generative models yielding the same observations, and those algorithms cannot discriminate the true causal model from other equivalent generative models. Recent progress have shown that it's impossible to to recover latent causal variables without inductive biases both on models and data sets~\cite{locatello2019challenging,khemakhem2020variational}. This problem is known as identifiability of causal models. Existing works established identifiability results based on the independent component analysis (ICA)~\cite{tharwat2020independent}. The identifiability and uniqueness of linear ICA models have be well studied \cite{eriksson2004identifiability}. For nonlinear ICA model, researchers argue that the latent causal variables are unidentifiable without temporal structure~\cite{hyvarinen2019nonlinear}. Recent advances focus on extending the identifiability of linear ICA to non-linear ICA, using the nonstationary structure of time series or auxiliary variables \cite{hyvarinen2019nonlinear}. However, this line of work doesn't assume the causal relationship or generative process between latent variables and observed variables, which limits its use. How to correctly identify latent causal variables and structure is still a open problem. Current attempt makes strong condition on measurement model, noise type, or require nonstationary time series \cite{xie2020generalized,kivva2021learning,yao2021learning}, and those methods are only tested on synthetic dataset or simple scenarios. Thus, there is an opportunity to identify latent causal variables in complex system, especially in the scenario where people have good domain knowledge, which is more informative than auxiliary variables.

\subsection{Uncertainty Quantification}
Uncertainty estimation in EAM enables the quantification of uncertainty stemming from the model structure and input/output data and improves our understanding of different scientific processes and inherent entity characteristics. Uncertainty estimates can be used to establish the usability of an entity-aware model for operational decision-making in real worl applications~\cite{lutjens2021physically,perdikaris2016multiscale}. Finally, uncertainty quantification (UQ) methods also allow domain scientists to encode prior knowledge as model structure~\cite{lavin2021simulation} for robust generalization. Uncertainty can be introduced in EAM due to several sources. EAM methods may be simplification or approximation of the real-world physical systems leading to a model structure-based uncertainty. Second, imperfections, measurement errors, interpolation, or noise in entity characteristics can also lead to uncertainty in the known characteristics. Finally, more recently, there has also been a focus on estimating distributional uncertainty that arises because of differences in the data distribution between training and test set.

Existing UQ methods include Bayesian methods that compute posterior prediction distribution and provide uncertainty estimates. Dropout-based methods like Monte Carlo Dropout~\cite{gal2016dropout} are utilized during the testing period for approximate Bayesian inference when making predictions. Weight perturbation schemes~\cite{mobiny2021dropconnect} have been adopted for weight-perturbation-based uncertainty quantification. Using variational inference makes learning in these Bayesian networks more feasible~\cite{blundell2015weight}. Other approaches, such as Mixture density networks~\cite{bishop1994mixture}, have been used for multi-modal data where each of the modalities can be captured using the mixing components. More comparisons of uncertainty estimation methods can be found in~\cite{psaros2022uncertainty}. Recent studies have also attempted to decompose different sources of uncertainties~\cite{wenger2022posterior}. Principles of evidential theory have further been used to learn other sources of uncertainty~\cite{sensoy2018evidential}.

Several of these UQ methods can be used to improve EAM methods discussed in Section~\ref{sec:methods}. First, several variational Gaussian processes methods~\cite{garnelo2018neural} use inducing points to estimate posterior function from few-shot data. Thus the uncertainty due to the use of different approximation mechanisms and different subsampled datasets can be estimated and used to study the difference in generalization capabilities of these methods. Second, the decomposition of uncertainty estimates can be pivotal in decision-making - understanding if the current EAM can help adapt the model to specific use cases or determine if we need to build better models and use different datasets for our analysis. Third, most UQ methods develop Bayesian frameworks that use Gaussian distribution as function priors. A direction that would be useful for practical applications is looking at other prior distributions for model parameter sampling. For instance, where a target or outcome variable (e.g., extreme temperature modeling) can take extreme values, approximating the prediction function using a Gumbel or t-distribution prior can enable more accuracy. Finally, existing EAM methods consider that all entities are independent. However, in many scenarios, the entity can also be a mixture of base entities, such as a community of people or a category of micro-organisms. While several multi-modal EAM methods exist, formulating the prediction function as a mixture of components also allows for multi-modal modeling.

\subsection{Fairness}
\label{sec:fairness}
In EAM, the imbalance in training data collected from multiple entities can naturally introduce bias for some entities or groups. Such entity-related bias can adversely affect both individual's opportunities and the inequity over the whole population. Another source of unfairness could be bias in measurement error of input features. For example, phenomena of datasets having higher error profile in emerging economies occurs in several applications~\cite{world2003improving,koumamba2021health}. Fairness over multiple entities can be commonly formulated in three different ways. First, individual fairness follows the philosophy that similar entities should yield similar predictions with respect to a particular task, regardless of sensitive attributes (e.g., gender, income, and race). The second type of fairness (e.g., equal opportunity~\cite{hardt2016equality} and statistical parity~\cite{dwork2012fairness}) aims to ensure that the model output distribution is fair across entities. Third, fairness can also be measured in terms of performance disparity across different entities, especially to identify biased predictions for entities in disadvantaged groups or low-resource environments. For these settings, fairness can also be defined over groups of entities formed by certain attributes. For example, fair flow prediction amongst river-streams groups that are grouped according to the local information of annual income and business type can reduce the chance of flood risks being underestimated for low-income areas.

Amongst existing fairness-enforcing methods, the most common strategy is to include additional fairness-related losses during the training process~\cite{serna2020sensitiveloss}. Another major direction is to learn group-invariant features~\cite{alasadi2019toward}, in which discriminators are introduced to penalize learned features with the discriminative information of certain sensitive attributes (e.g., gender). Sensitive category de-correlation also employs the adversarial learning regime, but it tries to mitigate the polarization of predictions~\cite{sweeney2020reducing,alasadi2019toward}. To alleviate the competition between the predictive accuracy and fairness, a bi-level model refinement is proposed to disentangle model prediction and fairness objective~\cite{xie2022fairness}. Another benefit of this method is that it allows non-differentiable fairness measures. On the other hand, new data collection and filtering methods are developed to reduce bias in downstream learning tasks \cite{jo2020lessons}. These methods have been applied to tasks related to face detection~\cite{serna2020sensitiveloss}, text analysis \cite{sweeney2020reducing}, land cover mapping~\cite{xie2022fairness}, etc.

Existing fairness-enforcing methods in EAM face several challenges. First, although many definitions of fairness have been proposed in existing literature, fairness needs to be carefully formulated depending on the nature of the target problems. Second,  fairness metrics are fragile or sensitive to the grouping of entities, i.e., conclusions on ``fair" or ``unfair" can be easily altered by simple changes grouping of entities. Third, in real-world EAM problems, the deployment environments may differ from the training environment. As a result, a fairness-enforced model learned from training samples may fail to preserve fairness in target testing scenarios.


\vspace{-0.1in}
\section{Conclusion}
In this survey, we proposed a structured review of entity-aware modelling (EAM) research. As shown by this paper, many different research efforts have the potential to advance EAM. We organized the existing research based on the availability of entity characteristics and training samples. We hope that this structure will help in providing an organized view of this rapidly evolving field of research. This survey will also be valuable for domain scientists interested in exploring the use of ML to enhance EAM in their respective applications. Furthermore, we presented additional research directions that will improve the performance and usability of EAM in operational decision-making.


\bibliographystyle{named}
{\footnotesize\bibliography{IJCAI/main}}

\end{document}